%% file: icmi2026_main.tex
\documentclass[sigconf]{acmart}
\newcommand{\blue}[1]{\textcolor{black}{#1}}
\usepackage{multirow}
\usepackage{comment}
\AtBeginDocument{%
  }

\copyrightyear{2026}
\acmYear{2026}
\setcopyright{cc}
\setcctype{by}
\acmConference[ICMI '26]{INTERNATIONAL CONFERENCE ON MULTIMODAL INTERACTION}{October 05--09, 2026}{Napoli, Italy}
\acmBooktitle{INTERNATIONAL CONFERENCE ON MULTIMODAL INTERACTION (ICMI '26), October 05--09, 2026, Napoli, Italy}
\acmDOI{10.1145/3776574.3831235}
\acmISBN{979-8-4007-2318-6/2026/10}

\begin{document}




\title{How Reliable Are Multimodal Signals of Conversational State?
  Evidence from \blue{Remote} Dyadic Collaborative Tasks}

\author{Tahiya Chowdhury}
\orcid{0000-0001-7219-2150}
\affiliation{
\institution{Colby College}
\state{Waterville}
\city{Maine}
\country{United States}}
\email{tahiya.chowdhury@colby.edu}

\renewcommand{\shortauthors}{Chowdhury}


\begin{abstract}
Measuring conversational states such as cognitive load and
conversational power from multimodal behavior requires characteristic features that are not only predictive but also reliable across task contexts.
We present a three-dimensional evaluation framework assessing
predictive accuracy, cross-task generalizability, and test-retest
reliability, applied to interactional, acoustic, and linguistic
features extracted from dyadic conversations during collaborative tasks performed over a video-conferencing \blue{platform}
(\mbox{AVCAffe} dataset; $53$ dyads, 9 tasks).
Our results show that no single feature family dominates all three
dimensions. Linguistic features show the highest predictive accuracy for cognitive
load but collapse under cross-task
evaluation, revealing sensitivity to task-specific vocabulary rather
than generalizable load signals.
Additionally, acoustic reliability, often reported as evidence of
feature stability, degrades once speaker identity is controlled, confirming that standard prosodic features measure vocal characteristics rather than conversational state.
Interaction features provide the only genuinely reliable signal
(unchanged after normalization, structurally immune to speaker
identity inflation), and floor dominance predicts within-dyad mental
load asymmetry.
Interestingly, \blue{classifying power role} remained near chance baseline
across all conditions, indicating limitations of task-level
aggregated behavior for predicting power role in conversation.
\blue{Our findings reveal three insights: (1) linguistic features predict best but generalize poorly across task contexts; (2) acoustic reliability collapses to near-zero once speaker identity is
controlled, challenging standard evaluation practice; and (3)
interaction features provide the only genuinely reliable signal,
with floor dominance predicting within-dyad cognitive load
asymmetry.} These results argue for speaker normalization and
multi-dimensional evaluation as prerequisites for context-aware, robust multimodal feature selection in conversational systems.

\end{abstract}

\maketitle


\section{Introduction}

Socially aware conversational systems, designed to support
collaboration, negotiation, and group decision-making, depend on
reliable measurement of latent conversational states such as cognitive
load and conversational power from observable multimodal behavior.
A growing body of work has explored acoustic prosody, turn-taking
dynamics, and linguistic content as informative signals for this
measurement task~\cite{vinciarelli2009social, schuller2013interspeech,
vukovic2021cognitive}, yet two critical questions remain largely
unaddressed: \emph{do these signals generalize across task contexts?}
and \emph{do they measure what they appear to measure, or do they
reflect other stable characteristics of speakers?}

Predictive accuracy, the dominant evaluation criterion in
multimodal affective computing, does not answer either question.
A feature may predict cognitive load well within a particular task
distribution while completely failing on novel task types.
Its reliability estimate may be inflated by stable speaker
characteristics (vocal tract anatomy, speaking style) rather than
genuine behavioral consistency during the conversation.
These distinctions matter enormously for system design: a feature
family that is predictive but not generalizable or reliable cannot
serve as the foundation of a real-time measurement system deployed
across diverse contexts.

In this paper, we examine predictive accuracy, reliability, and generalizability, all three dimensions simultaneously.
We evaluate interactional, acoustic, and linguistic feature families
using: (1) leave-one-dyad-out (LODO) cross-validation for predictive
accuracy; (2) leave-one-task category-out (LOCO) analysis across different task
types for generalizability; and (3) intraclass correlation
coefficients (ICC) before and after within-speaker normalization for
reliability.
We apply this three-dimensional framework to the AVCAffe dataset of
remote dyadic collaboration~\cite{sarkar2023avcaffe}, which provides
naturalistic Zoom interactions annotated for cognitive load and
conversational power across nine distinct tasks. \blue{Specifically, we train Random Forest models on task-level
behavioral features to predict self-reported cognitive load 
scores \footnote{Mental and temporal demand are used as measures for
cognitive load construct throughout this paper; they are two of the six NASA-TLX
subscales~\cite{hart1988nasa} selected for their moderate within-dyad agreement and sensitivity to
task-level conversational behavior in dyadic remote work (Section~\ref{data}).} (regression) and power class
(classification), evaluated under leave-one-dyad-out
cross-validation using Lin's concordance correlation
coefficient (CCC) and macro F1 respectively.} 

\blue{We organize our evaluation around four research questions:
\begin{itemize}
\item \textbf{RQ1}: Which feature family best predicts cognitive load
and conversational power under leave-one-dyad-out
cross-validation?
\item \textbf{RQ2}: Do predictive features generalize across task
categories?
\item \textbf{RQ3}: How reliable are features across repeated
instances of similar tasks, and does reliability change
after speaker normalization?
\item \textbf{RQ4}: Does floor dominance predict within-dyad
cognitive load asymmetry?
\end{itemize}
}

Our main contributions are:
\begin{itemize}
  \item A three-dimensional evaluation framework (prediction,
    generalizability, reliability) for multimodal feature selection that can be reused beyond the dataset used in this study.
  \item Evidence that linguistic features predict cognitive load
    significantly better than interaction or acoustic features,
    but are
    strongly task-dependent.
    \item Evidence that raw acoustic features in dyadic remote conversation reflect speaker identity rather than behavioral state: standard within-speaker normalization eliminates all apparent reliability, 
    raising concerns about inflated observations in multimodal affective computing applications.
  \item Floor dominance, an interaction feature measuring which one speaker controls the conversational floor, is significantly associated with load asymmetry between speakers within a dyad. 
\end{itemize}

\begin{figure}[t]
  \includegraphics[width=\columnwidth]{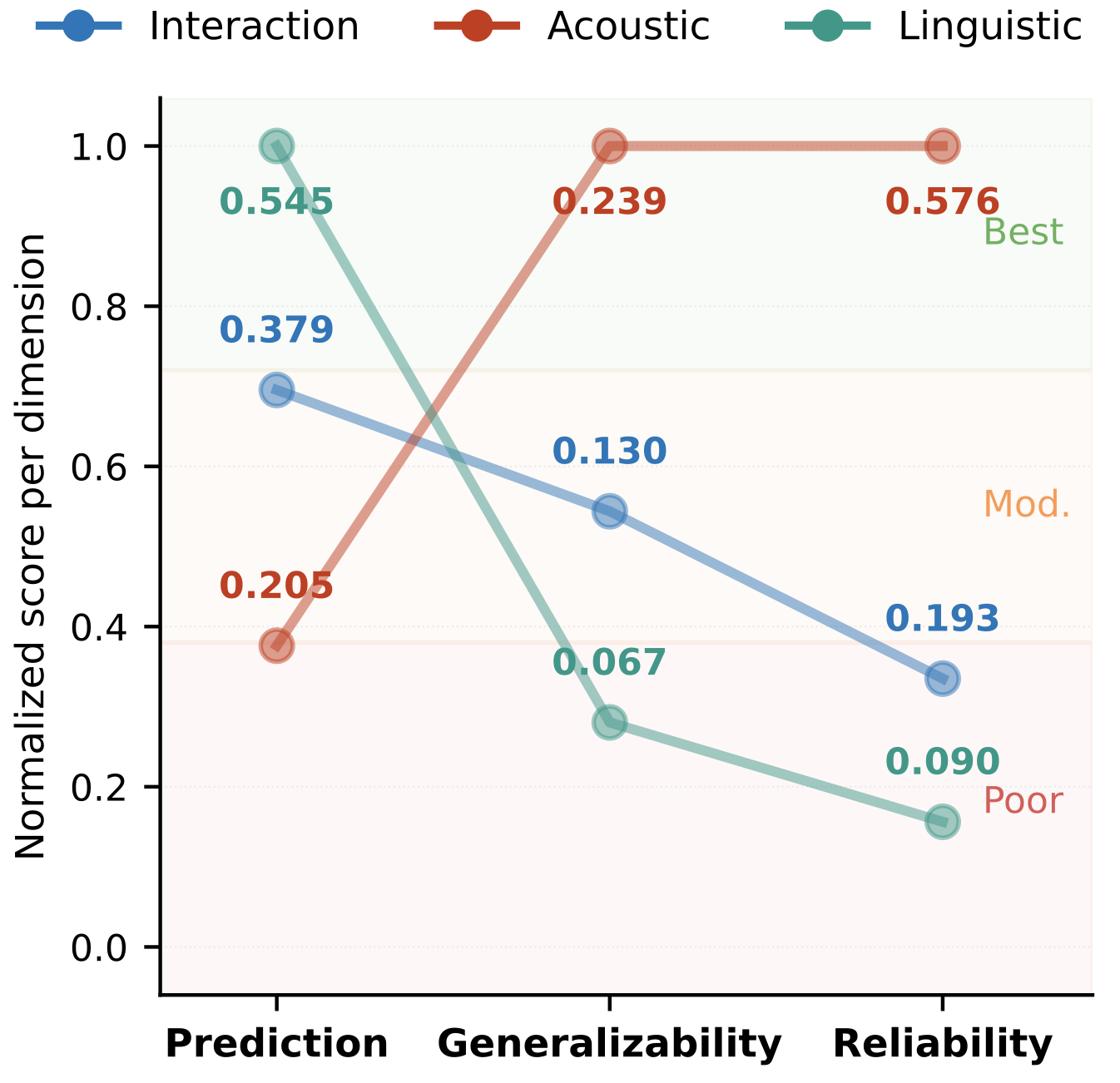}
  \caption{Our three-dimensional evaluation framework. Each line traces
    one feature family across prediction accuracy (mean LODO CCC), cross-task generalizability (mean LOCO CCC),
    and test-retest reliability (mean ICC(2,1)). Crossing lines reveal
    a rank dissociation that no feature family dominates all three dimensions.
    Background shaded zones indicate ICC reliability
    thresholds based on~\cite{koo2016icc}.}
    \Description{Our three-dimensional evaluation framework. Each line traces
    one feature family across prediction accuracy (mean LODO CCC), cross-task generalizability (mean LOCO CCC),
    and test-retest reliability (mean ICC(2,1)). Crossing lines reveal
    a rank dissociation that no feature family dominates all three dimensions.
    Background shaded zones indicate ICC reliability
    thresholds based on~\cite{koo2016icc}.}
  \label{fig:framework}
\end{figure}


Our findings draw attention to the need for context awareness in social interaction: features trained on one conversational context must generalize to others if they are to support context-aware interaction systems.



\section{Related Work}

\subsection{Speech Features for Cognitive Load}

Acoustic features have been the dominant modality for cognitive load
detection from speech.
Early work by Yin et al.~\cite{Yin2007, Yin2008} demonstrated that
pitch, speaking rate, and energy carry systematic load signatures.
The Interspeech 2014 Computational Paralinguistics
Challenge~\cite{Schuller2014} established speech-based cognitive and
physical load estimation as a benchmark problem, spurring a line of
work on acoustic features for cognitive load estimation in operational settings including aviation
communication~\cite{vukovic2021cognitive, yang2023cognitive} and
simulated flight~\cite{Xu2025}.
Boyer et al.~\cite{Boyer2018} and Taptiklis et al.~\cite{Taptiklis2023}
confirmed prosodic vocal biomarkers as reliable mental effort
indicators in controlled conditions.
However, this body of work evaluates acoustic features in single-task,
within-corpus settings; cross-task generalizability and speaker-level
reliability are not assessed.
\blue{To the best of our knowledge, no prior work has evaluated acoustic reliability
after within-speaker normalization in remote collaboration tasks, which we address in this work.} 

\subsection{Linguistic and Multimodal Features}

Khawaja et al.~\cite{khawaja2012analysis} showed that linguistic
features including hedges, fillers, and pronoun rates reflect
cognitive load in collaborative communication, directly motivating
linguistic feature family in our study.
Abel and Babel~\cite{abel2017cognitive} found that cognitive load
reduces linguistic convergence in dyads, suggesting that lexical
diversity is a theoretically grounded load marker.
Zhou et al.~\cite{zhou2018multimodal} proposed a multimodal
cognitive load measurement model combining behavioral and
physiological signals, arguing that multimodal data fusion improves
model robustness over any single data modality.
These findings motivate our multimodal feature family comparison. However, none of
these prior works has evaluated cross-task generalizability or
speaker-normalized reliability for cognitive load prediction from multimodal data, which is what we focus on in this work.

\subsection{Conversational Dynamics and Interaction Features}

Turn-taking patterns, overlap, and floor control have been studied
as indicators of social power and conversational
dominance in several prior works~\cite{vinciarelli2009social, gravano2011turn}.
Additionally, silence and response latency have been linked to cognitive
disengagement and processing time~\cite{heldner2010pauses,
johnstone1995there} indicating unequal participation in interaction as an indicator for cognitive load.
These interaction features motivate our dyad-level feature family to capture the conversation dynamics,
as their reliability and cross-task stability relative to commonly used acoustic
and linguistic families has not previously been evaluated.

\subsection{Datasets for Remote Collaborative Cognitive Load}

Sarkar et al.~\cite{sarkar2023avcaffe} introduced AVCAffe, the
first audio-visual dataset combining cognitive load and affect
annotations for remote work, which we use in this paper.
Their evaluation uses binary classification on 2-second video clips
with deep learning backbones, reporting weighted F1 (best: 65--67\%
for mental and temporal demand on long videos).
Our work differs fundamentally from their approach: we use task-level feature summaries consistent with the inherent nature of NASA-TLX~\cite{hart2006nasa, hart1988nasa}, employ
regression with Concordance Correlation Co-efficient (CCC) as our evaluation metric, and focus on measurement
validity in this work rather than model benchmarking based on performance only.

The recently introduced CoAffinity dataset~\cite{gunasekaran2025coaffinity}
provides a complementary resource for similar tasks: 39 participants performing eight remote-work tasks with audio, video, and physiological signals
(PPG, GSR) and cognitive load annotations.
Their results confirm that integrating physiological modalities
substantially improves load detection, suggesting that pure behavioral
signals face an inherent ceiling, a finding consistent with
our moderate CCC results across all three feature families.

\subsection{Reliability and Measurement Validity in Behavioral Signals}

We use the standard Intraclass Correlation Co-efficient (ICC) threshold provided by Koo and Li~\cite{koo2016icc} as a statistical measure for reliability throughout our reliability analysis. Despite ICC being the standard tool for test-retest reliability in behavioral measurement, its application to multimodal affective computing is relatively rare. 
The systematic inflation of raw acoustic ICC by speaker-specific
vocal characteristics (e.g., vocal tract anatomy, stable speaking
style) is a concern we demonstrate empirically for the first
time in a dyadic remote conversation dataset.
In a similar work, Koenecke et al.~\cite{koenecke2020racial} demonstrate unequal ASR error rates across speaker backgrounds, which has direct implications
for the reliability of linguistic features derived from automatically generated
transcripts for datasets with diverse participant pools such as AVCAffe
(18 countries of origin) with varying accent, gender, and other speaker conditions.

\begin{table}[t!]
\centering
\caption{Our contributions positioned relative to prior related work.}
\label{tab:comparison}
\setlength{\tabcolsep}{3.5pt}
\small
\begin{tabular}{lcccc}
\toprule
\textbf{Work} &
\textbf{Prediction} &
\textbf{Generalizability} &
\textbf{Reliability} \\
\midrule
Khawaja et al.~\cite{khawaja2012analysis}    & \checkmark & -- & --  \\
Vukovic et al.~\cite{vukovic2021cognitive}   & \checkmark & -- & -- \\
Sarkar et al.~\cite{sarkar2023avcaffe}       & \checkmark & -- & -- \\
Gunasekaran et al.~\cite{gunasekaran2025coaffinity} & \checkmark & -- & --\\
\textbf{Ours}                                & \checkmark & \checkmark & \checkmark \\
\bottomrule
\end{tabular}
\end{table}

\subsection{Positioning Our Contribution}

Table~\ref{tab:comparison} summarizes key distinctions between
our work and the most closely related studies.
Our contribution is threefold: (1) we introduce a three-dimensional
evaluation framework (prediction, generalizability, reliability) that
no other work applies jointly; (2) we provide a
speaker-normalized ICC analysis of acoustic features in a remote
conversation dataset, revealing systematic inflation by speaker
identity; and \blue{(3) we examine whether linguistic features' high predictive accuracy
generalizes across task contexts — a question prior work has not
addressed and has direct implications for the design of future context aware conversational systems.}


\section{Data}\label{data}

\subsection{AVCAffe Dataset}

We use AVCAffe~\cite{sarkar2023avcaffe}, an audio-visual dataset of
remote dyadic collaboration conducted over Zoom.
The dataset consists of 106 participants (49\% male, 50\% female,
1\% non-binary; age 18--57; 18 countries of origin) organized into
53 dyads.
Each dyad completed a session of nine task instances spanning seven
task types: open discussion, lighten the mood (sharing jokes),
Diapix (spot-the-difference in images), Montclair map (spatial coordination;
completed twice), Lost at Sea (group collaborative decision making), reading comprehension
(two rounds with roles reversed), and multi-task (email writing with
interruptions from partners).
Tasks were designed to elicit varying levels of cognitive demand,
from low (social/open tasks) to high (time-pressured coordination).
Ethics approval was obtained from the General Research Ethics Board
at Queen's University, Canada.

The participant pool spans a wide range of professional backgrounds
(engineers, scientists, students, nurses, and lawyers among others)
and age groups (18--57 years), with approximately 60\% from North
America and the remainder from India, Iran, and 14 other countries,
making it one of the most geographically diverse remote work
datasets available for cognitive load research using audio-visual recordings.

\begin{figure}[t]
  \includegraphics[width=\columnwidth]{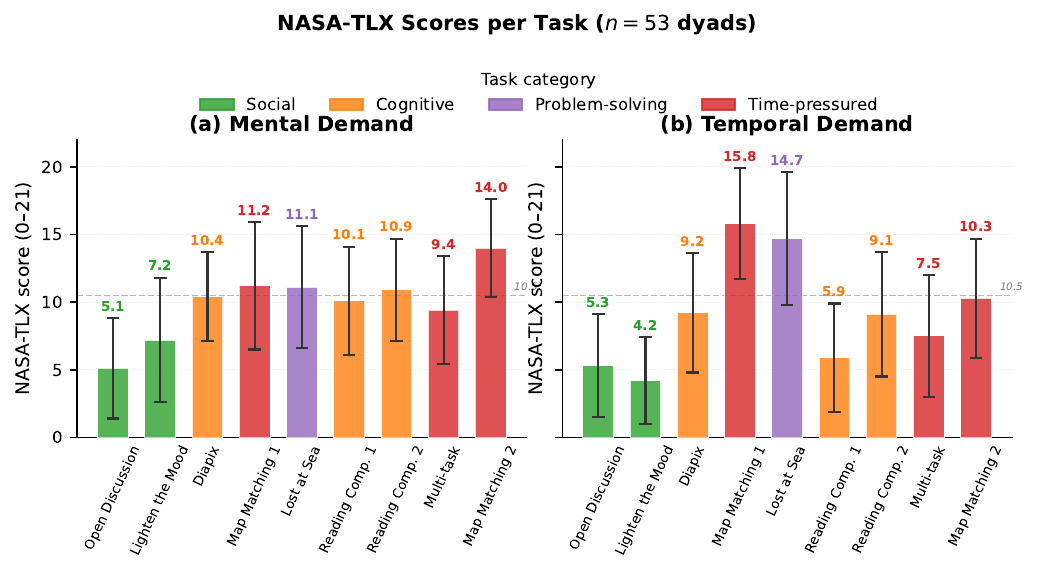}
  \caption{NASA-TLX cognitive load scores per task (mean~$\pm$~std,
    $n = 53$ dyads). Bar colors indicate task category.
    Social tasks (Tasks~1--2, green) elicit the lowest load on both
    dimensions, explaining why the Social category is the most
    challenging held-out condition in cross-task evaluation.
    Map Matching~1 (Task~4) and Lost at Sea (Task~5) show the highest
    temporal demand. The gray dashed line marks the scale midpoint (10.5).}
    \Description{NASA-TLX cognitive load scores per task (mean~$\pm$~std,
    $n = 53$ dyads). Bar colors indicate task category.
    Social tasks (Tasks~1--2, green) elicit the lowest load on both
    dimensions, explaining why the Social category is the most
    challenging held-out condition in cross-task evaluation.
    Map Matching~1 (Task~4) and Lost at Sea (Task~5) show the highest
    temporal demand. The gray dashed line marks the scale midpoint (10.5).}
  \label{fig:task_stats}
\end{figure}

\subsection{Annotations}

NASA-TLX ratings~\cite{hart1988nasa} were collected at the end of
each task on a 0--21 scale across six dimensions.
We use mental demand and temporal demand as primary regression
targets, selected for their moderate within-dyad agreement
($r = 0.341$ and $r = 0.300$ respectively) and theoretical relevance
to collaborative task engagement.
We exclude frustration ($r = 0.113$) and physical demand
($r = 0.071$) entirely as their low inter-rater agreement indicates these
dimensions reflect individual responses not systematically captured
by shared conversational behavior. Additionally, Effort showed redundant patterns with mental demand and is omitted from reported results.

\blue{\subsection{Prediction Targets}
\subsubsection{Cognitive Load.} For this regression task, we use the \textit{dyadic mean} of both speakers' rating as the cognitive load target, to be consistent
with the retrospective, shared-task nature of NASA-TLX annotation.}

\blue{\subsubsection{Conversational Power.} Power annotation was collected as a self-reported label per
participant per task across five categories (powerful, independent,
neutral, dependent, powerless). Due to high class imbalance,
we collapse these labels to three levels instead (high level: powerful +
independent; medium level: neutral; low level: dependent + powerless) yielding
$n = 354$ / $502$ / $94$ instances in the dataset.}

\subsection{Transcripts}

Speaker-level transcripts were generated using Distil-Whisper
ASR~\cite{gandhi2023distilwhisper} applied to the per-speaker,
per-task audio files provided as part of the dataset.
We note that AVCAffe participants represent 18 countries of origin.
As ASR transcription quality is known to vary across speaker
accents~\cite{koenecke2020racial, emara, mikel}, this may affect the reliability and accuracy
of linguistic features derived from these transcripts.


\section{Methods}

\subsection{Feature Families}

We evaluate three theoretically grounded feature families, all
computed at the task level (one value per speaker per task).

\textbf{Interaction features} (14 features, dyad-level) capture
conversational dynamics motivated by conversation analysis:
floor imbalance and turn count ratio~\cite{sacks1974simplest},
mean turn duration (both speakers), overlap ratio~\cite{gravano2011turn},
interruption imbalance and per-speaker interruption rates,
mean and standard deviation of response latency~\cite{levinson2015timing},
long silence duration ($\geq\!1$\,s), maximum inter-turn silence,
and dyadic dominance indices.
These features are identical for both speakers within a dyad-task
and are aggregated from 30-second window-level interaction features.

\textbf{Acoustic features} (10 features, per speaker) capture
prosodic and spectral characteristics motivated by load and arousal
research. This feature family includes pitch mean, std, and range; RMS loudness mean and std; voiced segments per minute; mean voiced segment duration; spectral centroid and bandwidth means; and estimated words per minute.

\textbf{Linguistic features} (21 features, per speaker) capture
speech content motivated by Linguistic Inquiry and Word Count (LIWC)~\cite{tausczik2010psychological}
and collaborative communication research~\cite{khawaja2012analysis}:
lexical diversity (type-token ratio and its root-normalized variant),
average word length, hedge and filler rates~\cite{khawaja2012analysis},
question and directive rates, first/second person pronoun rates and their ratio,
agreement and disagreement rates, backchannel and proposal rates,
modal verb, certainty, deference, negation, and politeness marker rates.

\blue{To summarize: interaction features are aggregated from 30-second windows to task level; acoustic and linguistic features are computed directly over the full task recording and full task transcript respectively, yielding one value
per speaker per task for all feature families. The resulting feature conditions range from 10 features (acoustic only) to 44 features (when all families combined) (see Table~\ref{tab:feature_comparison}).}

\subsection{Task-Level Analysis}

We analyze features at the task level to be consistent with the annotations provided with this dataset.
NASA-TLX is a retrospective self-report collected at the end of each
task, making the task the natural unit of analysis. While prior work has taken a window-level modeling approach by assigning
task-level labels to individual windows, this requires a Multiple Instance
Learning~\cite{pmlr-v80-ilse18a} assumption that we deliberately avoid here, keeping the measurement framework simple and interpretable.

\subsection{Evaluation Protocol}\label{sec:eval}

\blue{\textbf{Evaluation Metrics.} For cognitive load (regression task), we report Lin's concordance
correlation coefficient (CCC)~\cite{lin1989ccc}, which jointly
penalizes lack of correlation and systematic prediction bias.
For power (classification task), we report macro F1 with class weight
to compensate for class imbalance.
}

\subsubsection{\textbf{RQ1}: Leave One Dyad Out (LODO) for prediction accuracy.}
We use leave-one-dyad-out cross-validation over 53 folds for cross validation.
Features are standardized within each fold to prevent data leakage.
We use Random Forest (100 trees, \texttt{random\_state}=42) for
regression and classification~\cite{shwartz2022tabular}.

\subsubsection{\textbf{RQ2}: Leave One Category Out (LOCO) for cross-task generalizability.}
We apply leave-one-category-out cross-validation across four task
categories: Social/Open (tasks 1--2), Cognitive (tasks 3, 6, 7),
Problem-solving (task 5), and Time-pressured (tasks 4, 8, 9),
testing whether features trained on one task context generalize to
another. The tasks are grouped based on their nature and expected speaker interaction.

\subsubsection{\textbf{RQ3}: Intra-class Correlation Coefficient (ICC) for reliability.}
We compute ICC(2,1) --- two-way random effects, single rater,
absolute agreement~\cite{koo2016icc} on two test-retest pairs
within dyads: map-matching (tasks 4 and 9) and reading comprehension
(tasks 6 and 7), which share task type and demand level.
Our reliability thresholds follow Koo and Li~\cite{koo2016icc}:
ICC $> 0.75$ = good, $0.50$--$0.75$ = moderate, $< 0.50$ = poor.


\begin{table*}[t!]
\centering
\caption{%
  Leave-One-Dyad-Out (LODO) cross-validation ($n = 53$ dyads).
  CCC with 95\% bootstrap CI in brackets (2{,}000 samples;
  non-overlapping CIs indicate statistically supported differences).
  Power F1 = macro F1, 3-class; chance baseline\,=\,0.333.
  $\dagger$\,=\,best per column.
  Effort omitted (redundant with mental demand);
  performance omitted (negative CCC under cross-task evaluation).
}
\label{tab:feature_comparison}
\setlength{\tabcolsep}{4pt}
\renewcommand{\arraystretch}{1.2}
\begin{tabular}{l c ll ll ll}
\toprule
\multirow{2}{*}{\textbf{Feature Condition}} &
\multirow{2}{*}{\textbf{N}} &
\multicolumn{2}{c}{\textbf{Mental CCC}} &
\multicolumn{2}{c}{\textbf{Temporal CCC}} &
\multicolumn{2}{c}{\textbf{Power (F1)}} \\
\cmidrule(lr){3-4} \cmidrule(lr){5-6} \cmidrule(lr){7-8}
& & Mean & [95\,\% CI] & Mean & [95\,\% CI] & Mean & [95\,\% CI] \\
\midrule
Interaction only         & 14 &
  $+0.256$ & {\small[.184,.330]} &
  $+0.379$ & {\small[.309,.453]} &
  $0.329$  & {\small[.297,.363]} \\

Acoustic only            & 10 &
  $+0.184$ & {\small[.148,.220]} &
  $+0.205$ & {\small[.164,.247]} &
  $0.327$  & {\small[.293,.361]} \\

Linguistic only          & 21 &
  $+0.320$ & {\small[.260,.381]} &
  $+0.545^\dagger$ & {\small[.484,.605]} &
  $0.324$  & {\small[.290,.360]} \\

Interaction + Acoustic   & 24 &
  $+0.288$ & {\small[.228,.351]} &
  $+0.392$ & {\small[.325,.462]} &
  $0.310$  & {\small[.274,.348]} \\

Interaction + Linguistic & 35 &
  $+0.334$ & {\small[.268,.399]} &
  $+0.519$ & {\small[.458,.580]} &
  $0.359$  & {\small[.320,.400]} \\

All features             & 44 &
  $+0.343^\dagger$ & {\small[.280,.404]} &
  $+0.517$ & {\small[.458,.576]} &
  $0.338$  & {\small[.304,.375]} \\

\midrule
\textit{Chance baseline} & {---} &
  \multicolumn{2}{c}{---} &
  \multicolumn{2}{c}{---} &
  $0.333$  & {\small ---} \\
\bottomrule
\end{tabular}
\smallskip

\end{table*}

\subsubsection{\textbf{RQ4}: Load Asymmetry.} Floor dominance and within-dyad load asymmetry are analyzed using Pearson and Spearman correlation.

\textbf{Speaker normalization.}
To test whether acoustic (and linguistic) ICC is inflated by
speaker identity rather than behavioral consistency, we apply
within-speaker $z$-score normalization: for each speaker, each feature
is standardized across their nine task observations.
This removes between-speaker variance (vocal tract anatomy, stable
speaking style) while preserving within-speaker task-to-task
variation. 
\blue{Note that interaction features are dyadic relative measures (ratios and differences between speakers) that by construction cancel between-speaker absolute levels.} Negative ICC values after normalization indicate that within-speaker task-to-task variation is indistinguishable from residual error and no reliable signal remains once speaker identity is removed.

\begin{figure}[t]
  \includegraphics[width=\linewidth]{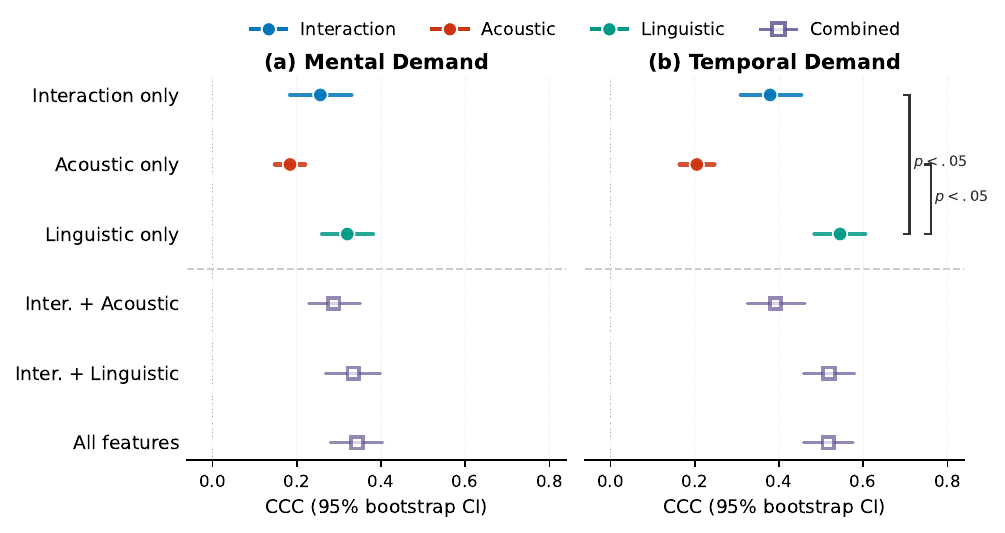}
  \caption{Predictive accuracy (LODO, $n=53$ dyads). Dots show mean
    CCC; lines show 95\% bootstrap CIs (2{,}000 samples). Filled
    circles = single-family conditions; open squares = combined.
    Brackets on panel~(b) indicate non-overlapping CIs
    ($p < .05$, bootstrap). CCC = Lin's concordance correlation
    coefficient~\cite{lin1989ccc}.}
    \Description{Predictive accuracy (LODO, $n=53$ dyads). Dots show mean
    CCC; lines show 95\% bootstrap CIs (2{,}000 samples). Filled
    circles = single-family conditions; open squares = combined.
    Brackets on panel~(b) indicate non-overlapping CIs
    ($p < .05$, bootstrap). CCC = Lin's concordance correlation
    coefficient~\cite{lin1989ccc}.}
  \label{fig:ccc}
\end{figure}

\begin{figure}[t]
  \includegraphics[width=\linewidth]{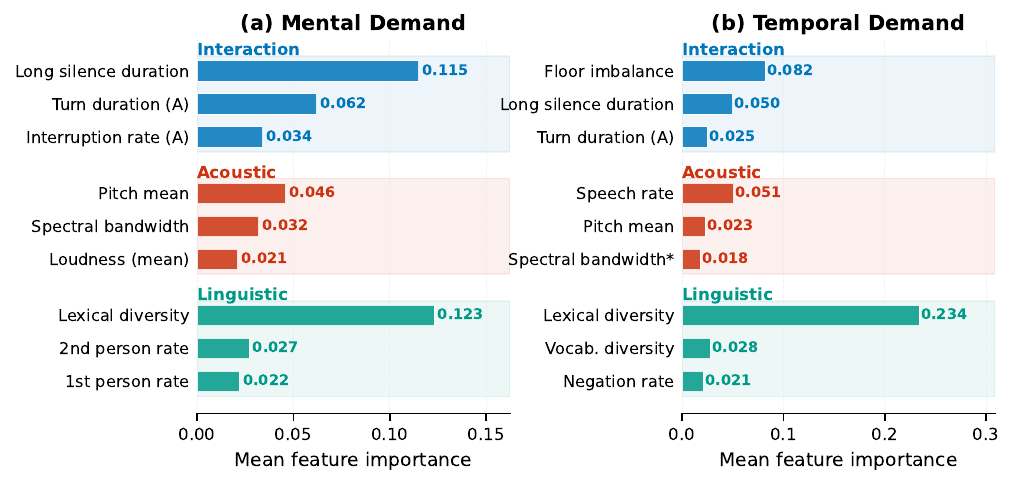}
  \caption{Top three features per family for each prediction target
  (Random Forest, all-features condition, LODO). Lexical diversity =
  root type-token ratio (proportion of unique words relative to square
  root of total words; higher diversity = more varied vocabulary).}
  \Description{Top three features per family for each prediction target
  (Random Forest, all-features condition, LODO). Lexical diversity =
  root type-token ratio (proportion of unique words relative to square
  root of total words; higher diversity = more varied vocabulary).}
  \label{fig:feature_imp}
\end{figure}

\section{Results}

\subsection{Predictive Accuracy (RQ1)}
\label{sec:results-pred}

\blue{\textbf{Statistical Criterion.} We report 95\% bootstrap confidence intervals (2{,}000 samples,
fold-level resampling) to support inferential interpretation of the regression and classification task given the small sample size. We use non-overlapping 95\% bootstrap CIs as our criterion for statistical support rather than a formal statistical test, given that no standard significance test exists for comparing LODO-fold CCC distributions with correlated folds~\cite{lin1989ccc}.}
Table~\ref{tab:feature_comparison} and Figure~\ref{fig:ccc} show
LODO cross-validation results for all feature conditions. 

\textbf{Linguistic features significantly outperform interaction and
acoustic families for temporal demand.}
Linguistic features achieve CCC\,=\,0.545, compared
to interaction (0.379) and acoustic
(0.205), with non-overlapping confidence intervals
confirming these differences are not attributable to sampling
variability.
Lexical diversity (root type-token ratio; importance\,=\,0.234 in the
all-features condition) is the dominant predictor for temporal demand,
consistent with the finding that vocabulary richness decreases as
speakers simplify language when under cognitive time
pressure~\cite{khawaja2012analysis, abel2017cognitive}.



\begin{table*}[t]
\centering
\caption{%
 LOCO cross-task generalizability: mental demand CCC when each
  task category is held out. Problem-solving contains one task ($n = 106$). \blue{Range column indicates difference between \textit{max}(CCC) and \textit{min}(CCC) across the four held-out categories per condition.}}

\label{tab:loco}
\setlength{\tabcolsep}{4pt}
\begin{tabular}{lc ccccc}
\toprule
\multirow{2}{*}{\textbf{Condition}} &
\multirow{2}{*}{\textbf{N}} &
\multicolumn{5}{c}{\textbf{Held-Out Task Category}} \\
\cmidrule(lr){3-7}
& & \textbf{Social} & \textbf{Cognitive} &
    \textbf{Problem} & \textbf{Time-pres.} & \textbf{Range}\\
\midrule
Interaction  & 14 & $+0.084$ & $+0.031$ & $+0.225$ & $+0.180$ & $0.194$\\
Acoustic     & 10 & $+0.121$ & $+0.240$ & $+0.355$ & $+0.238$ & $0.234$\\
Linguistic   & 21 & $+0.024$ & $+0.114$ & $+0.043$ & $+0.086$ & $0.090$\\
All features & 44 & $+0.153$ & $+0.154$ & $+0.234$ & $+0.220$ & $0.081$\\
\bottomrule
\end{tabular}
\smallskip

\end{table*}

\begin{figure*}[t!]
  \includegraphics[width=\linewidth]{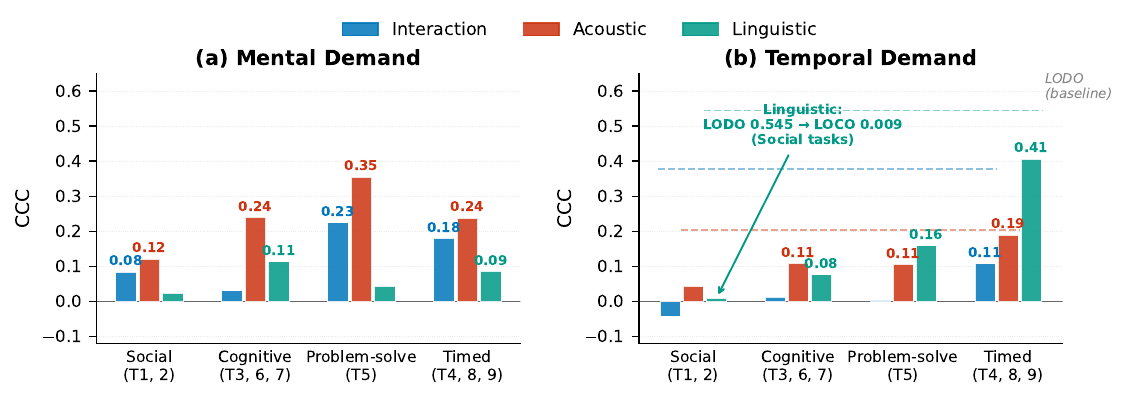}
  \caption{Cross-task generalizability results. Bars show CCC when the
    column category is held out for testing.
    Dashed lines in panel~(b) show within-distribution LODO performance as baseline.
    Linguistic temporal demand degrades from 0.545 (LODO) to
    0.009 on Social tasks.}
    \Description{Cross-task generalizability results. Bars show CCC when the
    column category is held out for testing.
    Dashed lines in panel~(b) show within-distribution LODO performance as baseline.
    Linguistic temporal demand degrades from 0.545 (LODO) to
    0.009 on Social tasks.}
  \label{fig:loco}
\end{figure*}

\textbf{For mental demand, differences between conditions are present but not statistically distinguished.}
All conditions produce CCC between 0.184 and 0.343, with broadly
overlapping confidence intervals, reflecting the limited statistical power of the 53-dyad sample.
Long silence duration (interaction importance\,=\,0.115) and lexical
diversity (linguistic importance\,=\,0.123) are the most informative
features for mental demand, consistent with silence as a
disengagement marker~\cite{heldner2010pauses} and with linguistic
load sensitivity~\cite{tausczik2010psychological}.

\textbf{Combining interaction and linguistic features does not
significantly improve over linguistic features alone.}
The interaction + linguistic condition (CCC\,=\,0.519 [0.458,\,0.580])
is statistically indistinguishable from linguistic only, suggesting
interaction features contribute complementary but not additive signal for temporal demand within this sample.

\textbf{Adding Acoustic features bring nominal value.}
Adding acoustic features to interaction improves temporal demand CCC by only 0.013, a trivial gain by combining multimodal information.

\textbf{Power classification is near chance.}
\blue{Conversational power is predicted as a 3-class problem
(high/neutral/low), evaluated with macro F1
under LODO evaluation protocol using a class-weighted Random Forest. The 3-class chance baseline (F1\,=\,0.333) falls within the 95\%
bootstrap CI for every feature condition (range: 0.310--0.359),
confirming that no feature family predicts conversational power above chance at the task level.}
We interpret this null result as reflecting the inherent subjectivity of power perception, label sparsity in the low-power class (9.9\%), and the limited temporal resolution in the data due to the task-level aggregation.

\subsection{Cross-Task Generalizability (RQ2)}
\label{sec:results-loco}

We present the results from cross-task generalizability experiments in Table~\ref{tab:loco} and Figure~\ref{fig:loco}.

\textbf{Linguistic features are strongly task-dependent.}
Temporal demand CCC drops from 0.545 (within-distribution, LODO) to
0.009 when tested on Social tasks \blue{(Table~\ref{tab:loco}, Linguistic row, Social column; Figure~\ref{fig:loco}b)}, while remaining high for
Time-pressured tasks (0.406).
This performance degradation reveals that linguistic features are learning
task-specific vocabulary patterns rather than a
generalizable load signal; high-demand tasks elicit
different speech content than social conversations.
The striking contrast between within-distribution and cross-task
performance (0.545 vs.\ 0.009) demonstrates that predictive
accuracy alone is insufficient as a measurement validity criterion.

\textbf{Acoustic features show more consistent cross-task performance.}
For mental demand, acoustic CCC ranges from 0.121 (Social) to 0.355
(Problem-solving), the most consistent profile across categories.
However, our results in the next section (Section~\ref{sec:results-icc}) show this consistency reflects speaker identity rather than genuine behavioral measurement.


\begin{table}[t]
\centering
\caption{%
  Test-retest ICC(2,1) before and after within-speaker z-score
  normalization; averaged over map-matching (tasks 4 \& 9) and
  reading comprehension (tasks 6 \& 7).
  Normalization removes between-speaker variance; negative
  normalized ICC\, means \,not reliable beyond speaker
  identity~\cite{koo2016icc}.
  Interaction features not normalized (dyadic relative measures).
  Thresholds: ICC\,$>$\,0.75\,=\,good; 0.50--0.75\,=\,moderate;
  $<$\,0.50\,=\,poor.
}
\label{tab:icc}
\setlength{\tabcolsep}{5pt}
\renewcommand{\arraystretch}{1.2}
\begin{tabular}{lc cc l}
\toprule
\multirow{2}{*}{\textbf{Family}} &
\multirow{2}{*}{\textbf{N}} &
\multicolumn{2}{c}{\textbf{Mean ICC(2,1)}} &
\multirow{2}{*}{\textbf{Interpretation}} \\
\cmidrule(lr){3-4}
& & \textbf{Raw} & \textbf{Normalized} & \\
\midrule
Interaction &
  14 & $+0.193$ \textit{\small(poor)} & $+0.193$ & Genuine \\[2pt]
Acoustic    &
  9 & $+0.576$ \textit{\small(mod.)} & $-0.112$ & Identity artifact \\[2pt]
Linguistic  &
  21 & $+0.090$ \textit{\small(poor)} & $-0.095$ & Speaker artifact \\
\bottomrule
\end{tabular}
\end{table}

\textbf{Combining features dampens task-dependence.}
The all-features condition (mental demand range: 0.153--0.234) is
more balanced than any single family, suggesting that combining
families reduces sensitivity to task-specific patterns.

We note that the Problem-solving category contains a single task
($n = 106$ test instances); estimates for this cell should be
interpreted with caution.

\subsection{Measurement Reliability (RQ3)}
\label{sec:results-icc}

\begin{figure}[t!]
  \includegraphics[width=\linewidth]{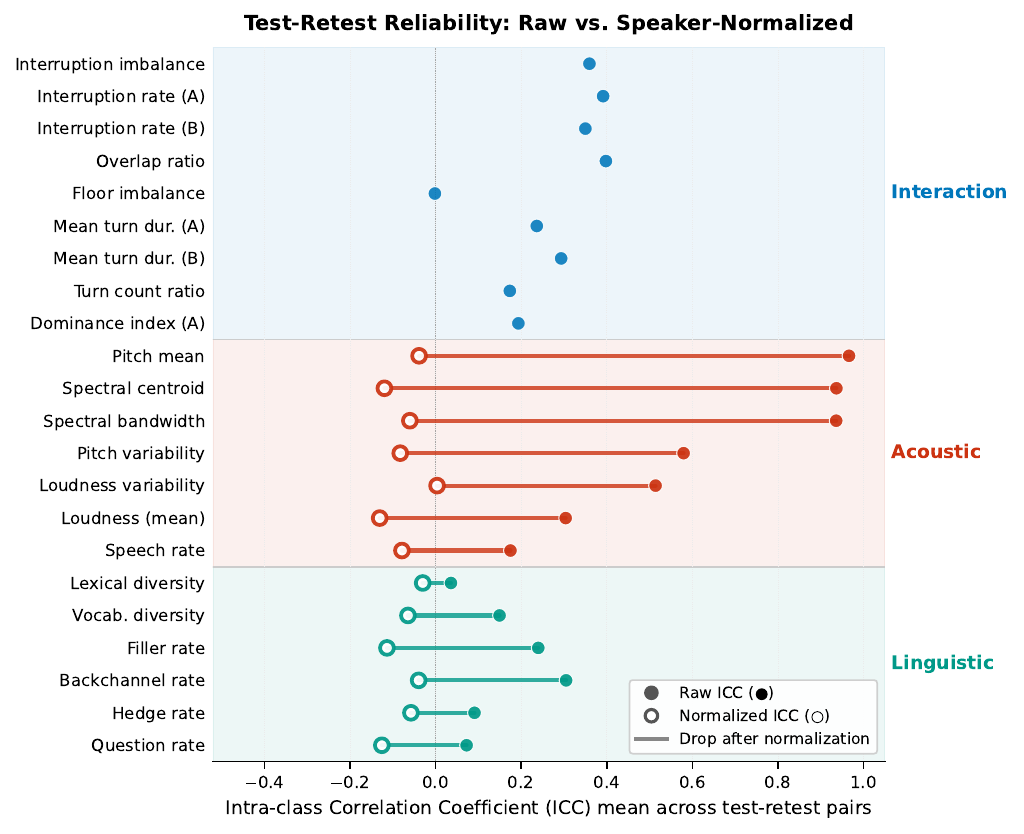}
  \caption{Test-retest reliability. Filled circles ({\Large\textbullet}) = raw ICC;
    open circles (\textopenbullet) = normalized ICC; lines = ICC drop after
    normalization. ICC(2,1) = two-way random effects, single rater,
    absolute agreement~\cite{koo2016icc}. Normalization removes
    between-speaker variance while preserving within-speaker variation.
    Acoustic features (red) show dramatic drops; interaction features
    (blue, single dots) remain unchanged.}
    \Description{Test-retest reliability. Filled circles ({\Large\textbullet}) = raw ICC;
    open circles (\textopenbullet) = normalized ICC; lines = ICC drop after
    normalization. ICC(2,1) = two-way random effects, single rater,
    absolute agreement~\cite{koo2016icc}. Normalization removes
    between-speaker variance while preserving within-speaker variation.
    Acoustic features (red) show dramatic drops; interaction features
    (blue, single dots) remain unchanged. Note that only representative features are shown here.} 
  \label{fig:icc}
\end{figure}

Table~\ref{tab:icc} and Figure~\ref{fig:icc} present ICC results
before and after within-speaker normalization.

\textbf{Acoustic ICC collapses entirely after speaker normalization.}
Raw acoustic ICC (mean: 0.576, moderate) drops to $-0.112$ after
within-speaker z-scoring, a 119\% reduction.
The most dramatic drops occur for spectral features: pitch mean
(0.966 $\to$ $-0.057$), spectral centroid (0.937 $\to$ $-0.178$),
and spectral bandwidth (0.936 $\to$ $-0.131$).
These are speaker-intrinsic properties reflecting vocal tract anatomy
rather than task-sensitive behavioral states.
Negative normalized ICC indicates that within-speaker variation across
tasks is indistinguishable from measurement error, no reliable
signal remains once speaker identity is removed.

\textbf{Linguistic ICC is largely artificial.}
Raw linguistic ICC (mean: 0.090) drops to $-0.095$ after
normalization.
Notably, lexical diversity (root type-token ratio), the most
predictive linguistic feature, has ICC\,=\,0.037 and 0.035 across
both test-retest pairs, near-zero even before normalization.
This confirms that lexical diversity is capturing task-specific
vocabulary rather than a stable individual characteristic.

\textbf{Interaction features provide genuine reliability.}
\blue{Table~\ref{tab:icc} confirms this observation empirically as Interaction
ICC is identical before and after normalization
($+0.193 \to +0.193$). Interaction ICC (mean: 0.193, poor) remains unchanged by normalization,
as dyadic relative measures (floor imbalance, interruption ratio) are
structurally immune to speaker identity inflation. }
While the overall level is poor by standard thresholds, it
represents genuine cross-task consistency post-normalization.

\textbf{Power label consistency is fair.}
Cohen's $\kappa$ between map-matching tasks 4 and 9 is 0.326 (fair),
and 0.237 between reading comprehension tasks 6 and 7, consistent
with the near-chance classification results.

\subsection{Load Asymmetry (RQ4)}

Within-dyad mental load asymmetry ($|\text{load}_A - \text{load}_B|$;
mean\,=\,5.57, SD\,=\,4.57) was significantly predicted by floor
imbalance ($r = 0.315$, 95\% CI [0.232,\,0.394], $p < 0.001$,
$n = 475$ dyad-task pairs).
Dyads with more unequal talk time distributions show greater
divergence in experienced cognitive burden.
No other interaction imbalance feature reached significance with
medium or greater effect size (correlation  $|r| < 0.12$,
Spearman $|\rho| < 0.08$).
These findings connect an observable behavioral asymmetry to the
distribution of subjective experience within dyads between speakers.


\section{Discussion}

\textbf{The three-dimensional framework as a diagnostic tool.}
Our results reveal a systematic dissociation across three
measurement dimensions.
Linguistic features are the best predictors but the worst in terms
of both generalizability and (normalized) reliability. They are
powerful within-distribution tools whose apparent reliability is an
artifact of between-speaker vocabulary differences and task-specific
content.
Acoustic features appear moderately reliable by conventional ICC but
this is entirely attributable to speaker-intrinsic spectral
characteristics.
Interaction features are genuine but modest in all three dimensions.
This suggests that no single family is suitable as a standalone measurement tool.\\

\subsection{Practical Implications}
We argue for three practical principles.

\begin {itemize}

\item First, \emph{speaker normalization is mandatory} before reporting
acoustic ICC: raw estimates are misleading and overestimate the
deployment value of prosodic features.

\item Second, \emph{cross-context evaluation} (our LOCO protocol) should
complement within-corpus cross-task validation to account for contextual differences; features that collapse
on held-out task types cannot support adaptive systems deployed
across conversational contexts.

\item Third, \emph{interaction features}, despite their moderate predictive
accuracy and poor ICC, are the most defensible foundation for
measurement systems because their reliability estimates are trustworthy, meaning
they generalize more consistently than linguistic features, and they
connect directly to observable behavioral events (floor control,
silence, interruption) during a conversation.
\end{itemize}

\blue{Note that our findings are grounded in Zoom-mediated dyadic
interaction and thus the results may differ across other media (e.g.,
in-person, audio-only) or multiparty interaction beyond dyadic settings. In multi-party
settings, power asymmetries and floor dynamics may be more pronounced, and acoustic normalization artifacts may vary
with recording equipment and platform-specific compression applied.}

\blue{\textbf{Implications for deployment.} The floor imbalance finding carries a concrete system design
implication: participation-aware conversational systems that monitor
the distribution of speaking time in real time could identify when one speaker is carrying disproportionate cognitive burden within a
dyad, enabling targeted support such as prompting the floor-dominant speaker to pause, or alerting a facilitator to rebalance participation.
Unlike acoustic features, floor imbalance is computable in real time from voice activity detection alone, without speaker
identification or normalization, making it a practical candidate
for deployment in low-latency conversational systems. Acoustic and linguistic features in their current form require offline ASR and spectral processing, and are not suitable for
real-time deployment without further engineering adaptation.}

\blue{\textbf{Power as a null finding.}
The failure to predict conversational power above chance from
any feature family, even combined, points to a fundamental
limitation of task-level behavioral aggregation for this target.
Three factors likely contribute to this.
First, the dyadic setup structurally constrains observable power
asymmetries: research on multi-party interaction has consistently
shown that dominance indices become more behaviorally evident
as group size increases, with speaking time asymmetries,
interruption patterns, and floor competition more discriminative
in group than dyadic
settings~\cite{vinciarelli2009social, schmidmast2002dominance,
danescu2012echoes}.
Second, computational work on power has found stronger signal
in structured multi-party settings such as institutional
discourse and online forums than in collaborative dyadic
conversation~\cite{danescu2012echoes}.
Third, power perception is subjective, dynamically negotiated
within conversations, and likely requires finer temporal
resolution and richer annotation schemes than task-level
self-reported class labels.}

\textbf{Recommendations.} For researchers working with existing multimodal datasets where speaker normalization was not applied during feature extraction, we recommend the following retrospective check: z-score each
acoustic feature within speaker across available task observations,
then recompute ICC(2,1) on the normalized values.
A drop exceeding 50\% relative to the raw ICC should be treated
as evidence that the reported reliability estimate reflects
speaker-intrinsic vocal characteristics rather than behavioral
consistency, and any system design claims based on that estimate
should be qualified and re-assessed accordingly.
This check requires no new data collection and can be applied to
any dataset where multiple task observations per speaker are available.

\subsection{Limitations}
This study has several limitations.
The sample is 53 dyads drawn from a single platform (Zoom) and
collected over a single study protocol. While the participant pool was diverse from different geographic regions and cultural context, the lack of systematic diversity may limit generalizability.

 The tasks used in this study appear in fixed order~\cite{sarkar2023avcaffe}, confounding
task type with session time, which can affect the implications of our findings.
Transcripts are generated by Distil-Whisper ASR. Due to ASR limitations, transcription
errors may be unequal across the 18 participant countries of origin,
potentially reducing the reliability of linguistic features for
speakers from underrepresented language
backgrounds~\cite{koenecke2020racial}.

For modeling purposes, we only evaluated Random Forest; other model families may show
different relative performance across feature families, which we left for future work.
LOCO estimates for the single-task Problem-solving category
($n = 106$) are unreliable due to the limited size of the dataset and should be replicated with larger
held-out sets.
 
\subsection{Future Work}
In this work we used dyad level mean as the target for cognitive load. A two-speaker latent state model tracking coupled individual
load trajectories with an explicit cross-speaker influence term is the natural extension of these findings and is the focus of
ongoing work.
Dynamical time series analysis techniques may capture coordination dynamics that is not accessible to
feature averages~\cite{fusaroli2012coming}, which we plan to explore as well.
Speaker-normalized acoustic reliability should be replicated on
other multimodal datasets to assess whether the identity inflation
we document is specific to Zoom or a more general artifact of
video-conferencing audio pipelines due to audio compression and other processing steps. \blue{Finally, these findings are based on remote dyadic setup, and we plan to extend this to multi-party setting as future work.}


\section{Conclusion}

We introduced a three-dimensional framework for evaluating
multimodal conversational features --- prediction accuracy,
cross-task generalizability, and speaker-normalized test-retest
reliability --- and applied it to dyadic Zoom-based collaboration.
Our central finding is that acoustic reliability is systematically
inflated by speaker identity: raw ICC values collapse to near-zero
after normalization, revealing that commonly used prosodic features
measure who speaks rather than how speakers behave under load.
Linguistic features predict best but generalize poorly.
Interaction features provide the most reliable signal.
These findings argue that speaker normalization and multi-dimensional
evaluation should be standard prerequisites for reliable feature selection
and model evaluation in multimodal conversational AI --- particularly
for systems intended to operate across diverse task contexts,
speaker populations, and conversational environments.


\section*{Responsible Innovation Statement}

This work uses the publicly released AVCAffe
dataset~\cite{sarkar2023avcaffe}, collected under ethics approval
from the General Research Ethics Board at Queen's University, Canada.
No new human data were collected as part of this study.
All analyses use anonymized participant identifiers.
We acknowledge that ASR-based transcription may introduce
systematic errors for speakers from underrepresented language
backgrounds~\cite{koenecke2020racial}, and we encourage future work
to validate linguistic features using manual or speaker-adapted
transcription.
The models and features evaluated here are not intended for
deployment without further reliability validation in target contexts.
Generative AI assistance was used for code and editorial
review during the preparation of this manuscript. The authors have meticulously reviewed all results and text and they take responsibility of all content of this paper.

\begin{acks}
We thank the AIIM lab, Queen's University, Canada, for collecting, preparing, and making this dataset publicly available. We also thank the anonymous reviewers for their helpful comments to improve this work. A special thanks to the students of CS466 Multimodal Interaction and Learning at Colby College (Spring 2026) for the discussion that motivated this work. This research was supported by the Henry Luce Foundation.
\end{acks}


\bibliographystyle{ACM-Reference-Format}
\bibliography{references}

\appendix
\input{appendix_features}

\end{document}

%% file: appendix_features.tex

\section*{Appendix: Feature Definitions}
\label{app:features}

Tables~\ref{tab:features_interaction}--\ref{tab:features_linguistic}
provide definitions for all 44 unique features used in this study,
organized by feature family.
All features are computed at the task level (one value per speaker
per task).
Interaction features are dyad-level (identical for both speakers
within a dyad-task); acoustic and linguistic features are
per-speaker.


\begin{table*}[ht]
\centering
\caption{Interaction features (14 features, dyad-level).
  Aggregated from 30-second window-level features across each task.
  These features are structurally relative --- they capture the
  asymmetry or joint behavior between the two speakers ---
  making them immune to speaker identity inflation in our analysis.}
\label{tab:features_interaction}
\setlength{\tabcolsep}{4pt}
\renewcommand{\arraystretch}{1.3}
\small
\begin{tabular}{p{3.8cm} p{6.2cm} p{5.2cm}}
\toprule
\textbf{Feature} & \textbf{Definition} & \textbf{Theoretical motivation} \\
\midrule

Floor imbalance &
  Normalized difference in total speaking time between Speaker A and
  Speaker B: $(t_A - t_B) / (t_A + t_B)$. Positive values indicate
  Speaker A dominates the floor. &
  Core measure of floor control~\cite{sacks1974simplest}.
  Predicts within-dyad cognitive load asymmetry in this study. \\

Turn count ratio (A) &
  Ratio of Speaker A's turn count to total turns taken by both
  speakers. &
  Captures who initiates and holds more conversational turns;
  linked to interactional
  dominance~\cite{vinciarelli2009social}. \\

Mean turn duration (A) &
  Average duration in seconds of Speaker A's speaking turns. &
  Longer turns indicate more processing time or narrative
  control~\cite{levinson2015timing}. \\

Mean turn duration (B) &
  Average duration in seconds of Speaker B's speaking turns. &
  Same as above for the other speaker; asymmetry between A and B
  is informative. \\

Overlap ratio &
  Proportion of task duration during which both speakers are
  simultaneously vocalizing. &
  Reflects conversational coordination and
  competition~\cite{gravano2011turn};
  high overlap may indicate engagement or conflict. \\

Interruption imbalance &
  Difference between Speaker A's interruption rate and
  Speaker B's interruption rate. Positive = A interrupts more. &
  Asymmetric interruption is a primary behavioral indicator of
  conversational
  dominance~\cite{vinciarelli2009social}. \\

Interruption rate (A) &
  Number of times Speaker A interrupts Speaker B per minute
  of task duration. &
  Per-speaker interruption reflects assertiveness and
  floor-taking behavior. \\

Interruption rate (B) &
  Number of times Speaker B interrupts Speaker A per minute
  of task duration. &
  Same as above for Speaker B. \\

Mean response latency &
  Mean time in seconds between one speaker's turn end and
  the other speaker's turn start, across all turn transitions. &
  Response latency reflects cognitive processing demands and
  coordination readiness~\cite{levinson2015timing}. \\

Std. response latency &
  Standard deviation of response latency across all turn
  transitions within the task. &
  High variability in latency reflects inconsistent coordination
  or fluctuating cognitive state. \\

Long silence duration &
  Total duration (seconds) of inter-turn silences
  $\geq 1.0$\,s within the task. &
  Prolonged silence marks cognitive disengagement or processing
  difficulty~\cite{heldner2010pauses}. \\

Max inter-turn silence &
  Longest single silence between consecutive turns, capped at
  10\,s to limit outlier influence. &
  Extreme silence events signal task breakdown or
  high cognitive load. \\

Dominance index (A) &
  Composite score for Speaker A combining floor share,
  interruption rate, and turn count ratio into a single
  dominance estimate. &
  Aggregate behavioral dominance
  measure~\cite{vinciarelli2009social}. \\

Dominance index (B) &
  Same composite dominance score for Speaker B. &
  Symmetric measure allowing A--B comparison. \\

\bottomrule
\end{tabular}
\end{table*}


\begin{table*}[ht]
\centering
\caption{Acoustic features (10 features, per speaker, task-level).
  Computed from per-speaker audio recordings using voice activity
  detection and pitch extraction.
  \textbf{Note:} As shown in Table~4 of the main paper, raw ICC
  values for spectral features (pitch mean, spectral centroid,
  spectral bandwidth) collapse to near-zero after within-speaker
  normalization, indicating they primarily reflect stable vocal
  tract characteristics rather than task-sensitive behavioral states.}
\label{tab:features_acoustic}
\setlength{\tabcolsep}{4pt}
\renewcommand{\arraystretch}{1.3}
\small
\begin{tabular}{p{3.8cm} p{6.2cm} p{5.2cm}}
\toprule
\textbf{Feature} & \textbf{Definition} & \textbf{Theoretical motivation} \\
\midrule

Pitch mean (Hz) &
  Mean fundamental frequency (F0) of voiced frames across
  the task, in Hertz. &
  Pitch elevation is associated with arousal, stress, and
  cognitive
  load~\cite{vukovic2021cognitive, Boyer2018};
  also reflects stable speaker vocal tract anatomy. \\

Pitch std (Hz) &
  Standard deviation of F0 across voiced frames. &
  Pitch variability reflects prosodic expressiveness;
  reduced variability under high load. \\

Pitch range (Hz) &
  Difference between maximum and minimum F0 across
  voiced frames. &
  Wide pitch range indicates more expressive speech;
  range narrows under cognitive
  demand~\cite{schuller2013interspeech}. \\

RMS loudness (mean) &
  Mean root-mean-square energy of the speech signal,
  a proxy for vocal loudness. &
  Loudness is associated with dominance and
  arousal~\cite{vinciarelli2009social};
  sensitive to Zoom AGC processing. \\

RMS loudness (std) &
  Standard deviation of RMS energy across the task. &
  Loudness variability reflects engagement and
  expressiveness; may reflect AGC instability in
  videoconferencing. \\

Voiced segments/min &
  Number of voiced speech segments per minute of task
  duration, estimated via voice activity detection. &
  Speaking rate proxy; cognitive load slows speech
  production~\cite{Yin2007}. \\

Mean voiced segment dur. &
  Average duration in seconds of continuous voiced
  speech segments. &
  Longer segments indicate more fluent, less hesitant
  speech; shorter segments indicate more
  fragmentation~\cite{khawaja2012analysis}. \\

Spectral centroid (mean) &
  Mean of the spectral centroid (frequency-weighted mean
  of the power spectrum) across voiced frames, in Hz. &
  Voice brightness proxy; reflects vocal tract resonance
  characteristics. High raw ICC confirmed to be a
  speaker identity artifact in this study. \\

Spectral bandwidth (mean) &
  Mean spectral bandwidth (spread of the power spectrum
  around the centroid) across voiced frames. &
  Reflects spectral shape of voice; stable across tasks
  due to vocal tract anatomy rather than behavioral state. \\

Estimated words/min &
  Estimated speaking rate derived from transcript word
  count divided by voiced duration. &
  Direct cognitive load indicator; speakers slow down
  under increased mental demand~\cite{Yin2008,
  Taptiklis2023}. \\

\bottomrule
\end{tabular}
\end{table*}


\begin{table*}[ht]
\centering
\caption{Linguistic features (21 features, per speaker, task-level).
  Extracted from per-speaker task-level transcripts generated by
  Distil-Whisper ASR~\cite{gandhi2023distilwhisper}.
  Rate features are normalized by total word count or segment count.
  }
\label{tab:features_linguistic}
\setlength{\tabcolsep}{4pt}
\renewcommand{\arraystretch}{1.3}
\small
\begin{tabular}{p{3.8cm} p{6.2cm} p{5.2cm}}
\toprule
\textbf{Feature} & \textbf{Definition} & \textbf{Theoretical motivation} \\
\midrule

Lexical diversity &
  Root type-token ratio (root TTR): number of unique word types
  divided by the square root of total word tokens. Higher values
  indicate more varied vocabulary. &
  Lexical diversity decreases under cognitive load as speakers
  simplify language~\cite{khawaja2012analysis,
  abel2017cognitive}. Top predictor of temporal demand in
  this study (importance $= 0.234$). \\

Vocab. diversity (alt.) &
  Standard type-token ratio: unique word types divided by total
  word tokens. Complementary to root TTR. &
  Alternative operationalization of lexical
  diversity~\cite{tausczik2010psychological}. \\

Avg. word length &
  Mean number of characters per word across the transcript. &
  Longer words indicate more complex vocabulary; word
  complexity decreases under high cognitive load. \\

Hedge rate &
  Proportion of words that are hedging expressions
  (e.g., \textit{maybe, perhaps, I think, sort of}). &
  Hedging signals epistemic uncertainty; increases under
  cognitive load as speakers express less
  confidence~\cite{khawaja2012analysis}. \\

Filler rate &
  Proportion of words that are filler expressions
  (e.g., \textit{uh, um, like, you know}). &
  Fillers mark disfluency; increase under cognitive load as
  speech production is disrupted~\cite{khawaja2012analysis}. \\

Question rate &
  Proportion of utterance segments ending with a
  question mark. &
  Question frequency reflects information-seeking and
  coordination strategies between speakers. \\

1st person singular rate &
  Proportion of words that are first-person singular
  pronouns (I, me, my, mine, myself). &
  Self-referential language reflects individual focus;
  decreases in collaborative framing~\cite{tausczik2010psychological}. \\

1st person plural rate &
  Proportion of words that are first-person plural pronouns
  (we, us, our, ours, ourselves). &
  Collective framing reflects collaborative orientation;
  increases in successful coordination. \\

2nd person rate &
  Proportion of words that are second-person pronouns
  (you, your, yours, yourself). &
  High second-person rate indicates direct addressee
  engagement and task direction. \\

Pronoun ratio (I/you) &
  Ratio of first-person singular to second-person pronoun
  usage. &
  Self vs.\ other orientation in collaborative dialogue;
  high values indicate more self-focused speech. \\

Agreement rate &
  Proportion of utterance segments containing agreement
  markers (e.g., \textit{yes, right, exactly, okay}). &
  Agreement frequency reflects collaborative alignment
  and conversational
  coordination~\cite{khawaja2012analysis}. \\

Disagreement rate &
  Proportion of utterance segments containing disagreement
  markers (e.g., \textit{no, but, actually, I disagree}). &
  Disagreement reflects cognitive friction and divergent
  problem-solving. \\

Backchannel rate &
  Proportion of short segments identified as backchannels
  (e.g., \textit{mm-hm, yeah, uh-huh}). &
  Backchannels signal active listening and turn-yielding
  behavior; reflect conversational engagement. \\

Directive rate &
  Proportion of utterance segments classified as directives
  (imperative or instruction-giving speech acts). &
  High directive rate is associated with leadership and
  conversational power~\cite{vinciarelli2009social}. \\

Proposal rate &
  Proportion of utterance segments classified as proposals
  (suggesting or initiating a joint action). &
  Proposals reflect collaborative planning and
  initiative-taking in task-oriented dialogue. \\

Modal verb rate &
  Proportion of words that are modal verbs
  (e.g., \textit{might, should, could, would, must}). &
  Modal verbs mark epistemic stance; high rates indicate
  tentative or conditional
  reasoning~\cite{tausczik2010psychological}. \\

Certainty marker rate &
  Proportion of words that are certainty markers
  (e.g., \textit{definitely, certainly, clearly, obviously}). &
  Certainty expression reflects speaker confidence and
  may decrease under cognitive load. \\

Deference rate &
  Proportion of utterance segments containing deference
  markers (e.g., \textit{please, sorry, excuse me,
  if you don't mind}). &
  Politeness and deference reflect power asymmetry and
  social orientation in conversation. \\

Negation rate &
  Proportion of words that are negation words
  (e.g., \textit{not, no, never, neither, nor}). &
  Negation frequency reflects disagreement, rejection,
  or correction in task-oriented dialogue. \\

Politeness rate &
  Proportion of utterance segments containing explicit
  politeness markers (e.g., \textit{thank you, please,
  you're welcome}). &
  Social coordination marker; reflects relationship
  management within the dyad. \\

Estimated words/min &
  Estimated speaking rate: total word count divided by
  voiced audio duration in minutes.
  (Also included in acoustic features.) &
  Lexical production rate; slows under cognitive load
  and captures speech-content load signal when combined
  with linguistic features. \\

\bottomrule
\end{tabular}
\end{table*}